\documentclass[twoside,11pt]{article}
\usepackage{jmlr2e}
\usepackage{amsmath}

\ShortHeadings{Vehicle predictive isochronous trajectory}{Damian D.}
\firstpageno{1}
\begin{document}
\title{Vehicle predictive trajectory patterns from isochronous data}
\author{\name Dumitru Damian \email dumitrudamian@yahoo.com \\
       \addr Engineering Consultant\\
       \addr Deva, 33012, RO}
\editor{open}
\maketitle
\begin{abstract}
Measuring and analyzing sensor data is the basic technique in vehicle dynamics development and with the advancement of embedded and data acquisition systems it is possible to analyze large data sets. In this paper a detailed method is presented for assessing and mapping isochronous trajectory patterns in Graz (Austria) by using data fusion from video, ArduinoUno and the compass sensor HDMM01. The predictive isochronous trajectory patterns are derived from the data values for a predefined time horizon. Both extreme driving behavior and hazardous road geometries can be identified. It is possible to provide instant road sensor data which can be used to compare the data from a trajectory path as well as for different time instances. Results of this study show that the trajectory patterns are successful in predicting the likely evolution of a current trajectory pattern and can provide assessment on future driving situations. The obtained data from this study can be useful as reference in future city planning for energy saving driving pathways as well as vehicle design and engineering improvements based on quantitative and relevant dynamic measurements.
\end{abstract}
\section{Introduction}
	Predicting vehicle behavior is a major challenge and an active research domain for the automotive industry. The desire to develop intelligent vehicles that make good use of predictive vehicle behavior and contribute to the improvement of road safety can find a possible implementation by using the low cost compass sensor data.

	Complex planning algorithms have begun to factor in uncertainty that arise from variations in travel time, erratic communication between autonomous vehicles, imperfect sensor data or other situations. To consider such scenarios, a method for path planning is developed that also generates contingency vehicle paths, should the initial path prove too risky. By using the isochronous trajectory patterns it is also possible to identify particular conditions that consider sensor readings or delays in the measured data, that must trigger a switch to a particular contingency vehicle path.
		
\emph {The isochronous trajectory is defined to be a sequence of position vectors in the 3-dimensional Euclidean space, the notation used is $\Gamma$ and has the following representation:}

 \[  \Gamma_{t} = (x_{t},y_{t},z_{t})\] where t=20ms.

	The coordinate system used in vehicle dynamics modeling is according to SAE J670e as shown in Figure 2. 

\begin{figure}[htbp]
   \begin{minipage}{0.6\textwidth}
     \centering
     \includegraphics[width=.8\linewidth]{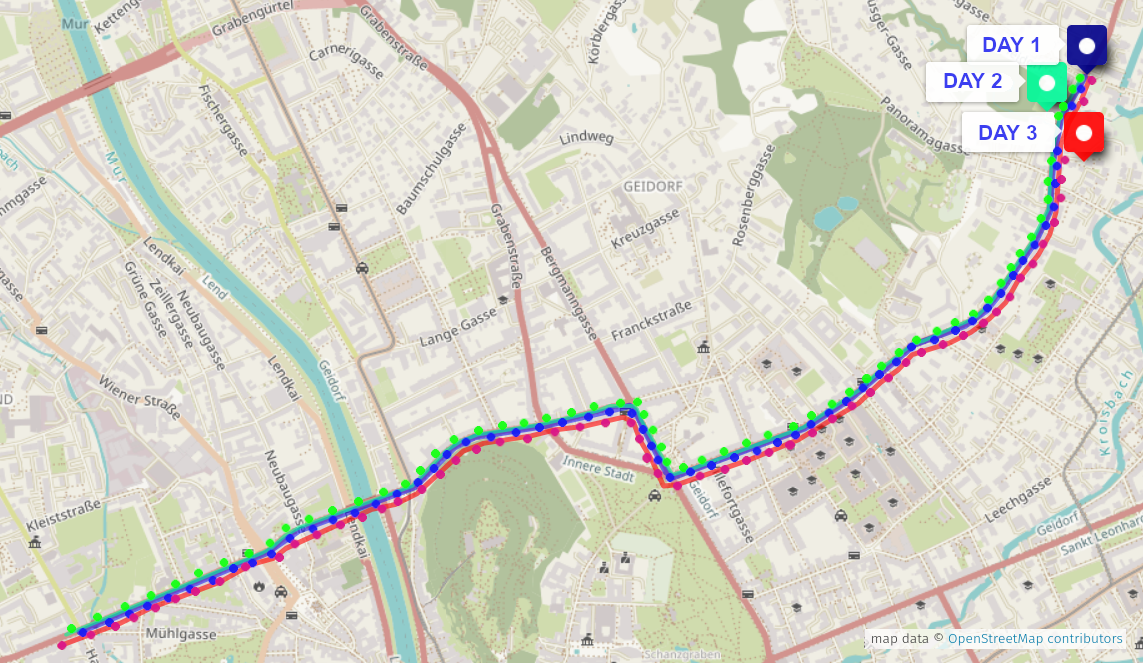}
     \caption{Vehicle path with isochronous data}\label{figure1}
   \end{minipage}\hfill
   \begin{minipage}{0.4\textwidth}
     \centering
     \includegraphics[width=.85\linewidth]{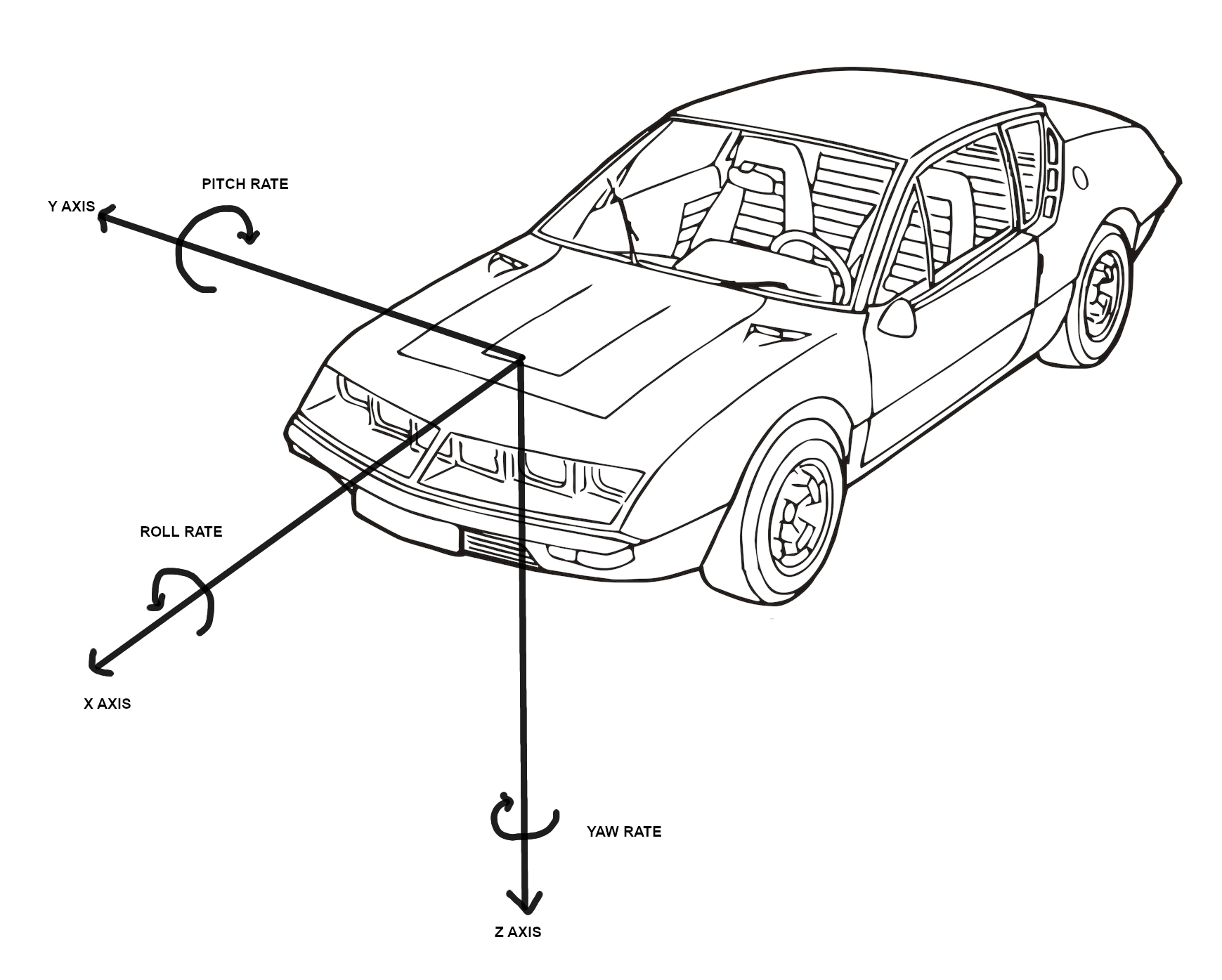}
     \caption{Vehicle axis system}\label{Vehicle axis}
   \end{minipage}
\end{figure}

	From the vehicle axis system it is possible to construct the isochronous surface. The X, Y and Z direction can be represented by one or multiple nodes that describe the isochronous surface. 
	
	The vehicle is considered to be a rigid body, the position vector for a point P in the XYZ coordinates system, will have the following representation:
\[ \vec r = \vec r_t  \]
\[ \vec r = x_t*\vec i +  y_t*\vec j + z_t*\vec k\]

\begin{figure}[htbp]
   \begin{minipage}{0.5\textwidth}
     \centering
     \includegraphics[width=.85\linewidth]{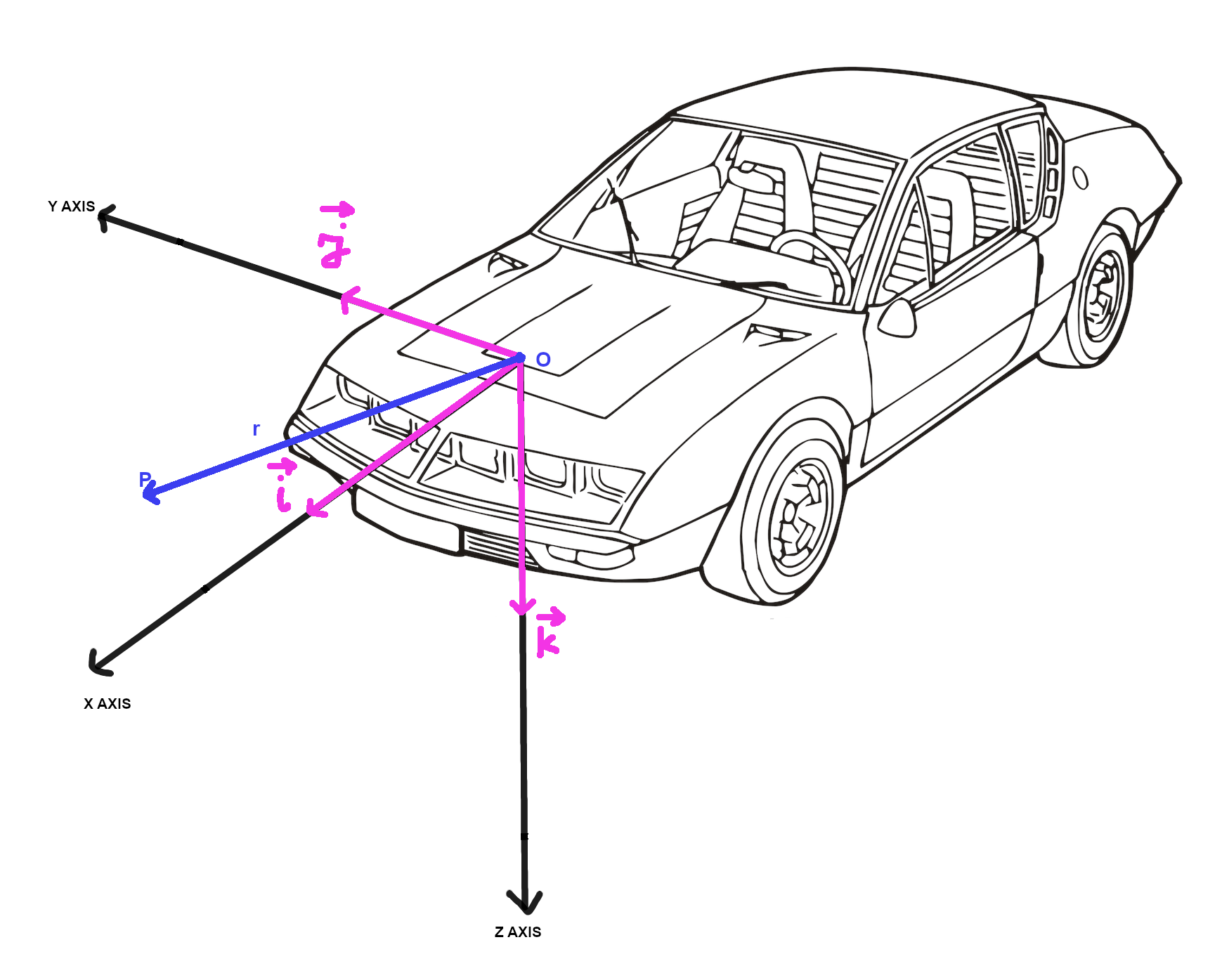}
     \caption{OP positional vector}\label{Fig:OP positional vector}
   \end{minipage}\hfill
   \begin{minipage}{0.5\textwidth}
     \centering
     \includegraphics[width=.85\linewidth]{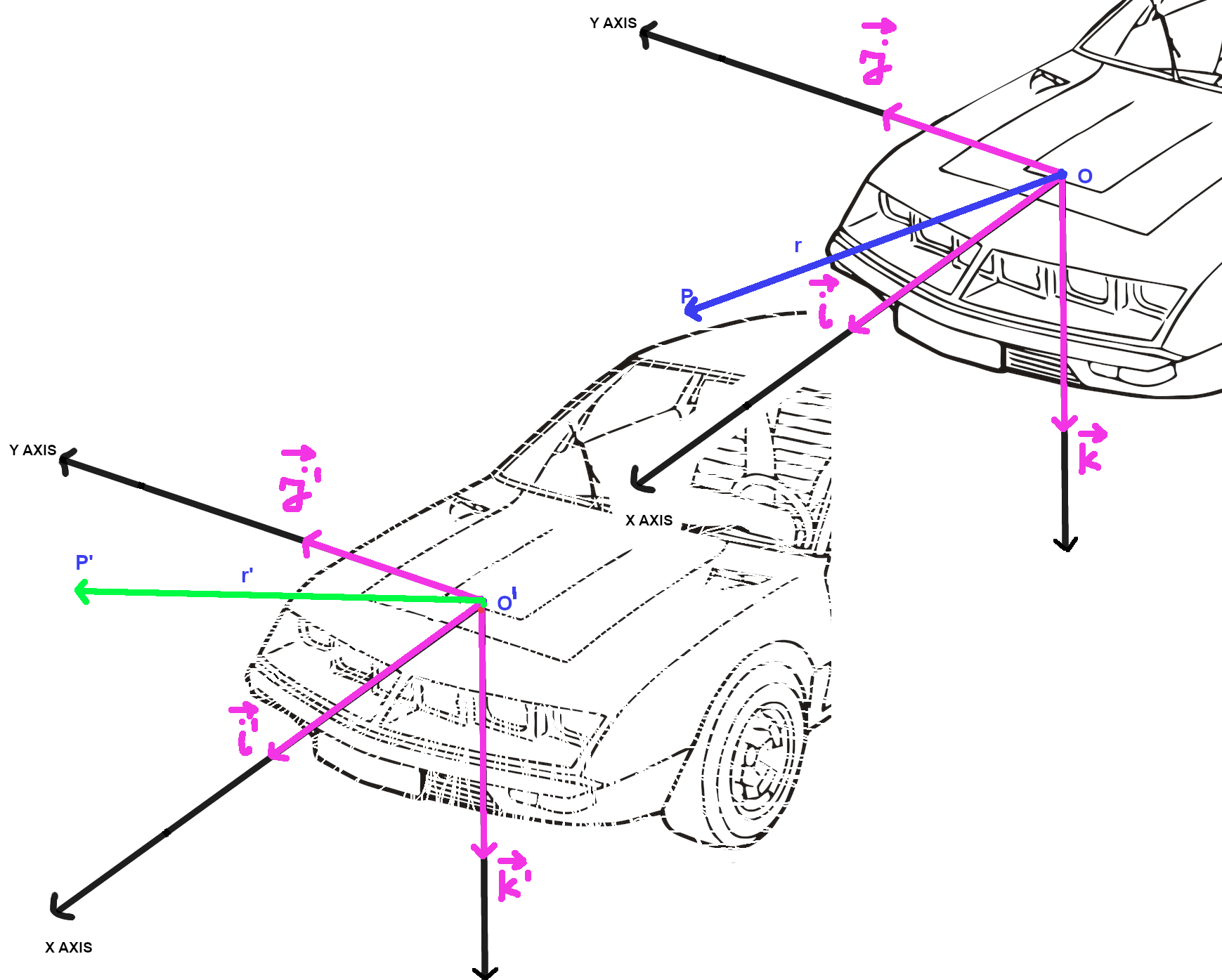}
     \caption{O'P' vector with new coordinates}\label{O'P' vector with new coordinates}
   \end{minipage}
\end{figure}

	The position vector for a point P' in the mobile X’Y’Z’ coordinates system can be localized trough: 
\[ \vec r' = \vec r'_t  \] 
\[ \vec r' = x'_t*\vec i' +  y'_t*\vec j' + z'_t*\vec k' \]

	Also, the origin O’ for the mobile X’Y’Z’ coordinates system in comparison with the origin O for the XYZ coordinates system can be represented trough the position vector:
\[ \vec r_0 = \vec r_{0_t} \]
\[ \vec r_0 = x_{0_t}*\vec i +  y_{0_t}*\vec j + z_{0_t}*\vec k \] 
and thus: 
\[ \vec r_t= \vec r_{0_t} + \vec r'_t \] 

The relation between $ \vec r'_t $ and $ \vec r_t $ can be represented trough the following relation:
\[ i' = \alpha_{11}*i + \alpha_{12}*j + \alpha_{13}*k \]
\[ j' = \alpha_{21}*i + \alpha_{22}*j + \alpha_{23}*k \]
\[ k' = \alpha_{31}*i + \alpha_{32}*j + \alpha_{33}*k \]

And from this dependency the isochronous matrix that characterizes the state change for one time step is determined:
\[
A = 
\begin{pmatrix}
\alpha_{11} & \alpha_{12} & \alpha_{13}\\
\alpha_{21} & \alpha_{22} & \alpha_{23}\\
\alpha_{31} & \alpha_{32} & \alpha_{33}
\end{pmatrix}
\] 

From the above representation it is possible to derive the velocity isochronous matrix:
\[ \vec v = \frac{d \vec r}{dt} = \dot{\vec{r}} = \frac{d \vec r_{0}}{d t} + \frac{d \vec r'}{d t} = \dot{\vec{r_0}} + \dot{\vec{r'}} \]
\[ \dot{\vec{r_0}} = \dot{x_{0_t}} \vec i +  \dot{y_{0_t}} \vec j + \dot{z_{0_t}} \vec k \]
\[ \dot{\vec{r'}} = \dot {x}' \vec{i'} + \dot{y}'\vec{j'}+ \dot{z'} \vec{k'} \]

Each isochronous surface can have multiple nodes that characterize the state of respective surface. For one mini-node, the vehicle velocity isochronous matrix can be represented:
\begin{equation}  \label{eq:isochronous matrice}
V = 
\begin{pmatrix}
\upsilon_{11} & \upsilon_{12} & \upsilon_{13}\\
\upsilon_{21} & \upsilon_{22} & \upsilon_{23}\\
\upsilon_{31} & \upsilon_{32} & \upsilon_{33}
\end{pmatrix}
\end{equation} 

The equation \eqref{eq:isochronous matrice} can be applied to each mini-node that forms the isochronous surface. For when there are 9 mini-nodes, the vehicle velocity isochronous matrix will be:
	
\begin{equation}  \label{eq:isochronous main node}
V_{iso} = 
\begin{pmatrix}
\begin{pmatrix} 
\upsilon_{111} & \upsilon_{112} & \upsilon_{113}\\
\upsilon_{121} & \upsilon_{122} & \upsilon_{123}\\
\upsilon_{131} & \upsilon_{132} & \upsilon_{133}
\end{pmatrix}  &  
\begin{pmatrix} 
\upsilon_{211} & \upsilon_{212} & \upsilon_{213}\\
\upsilon_{221} & \upsilon_{222} & \upsilon_{223}\\
\upsilon_{231} & \upsilon_{232} & \upsilon_{233}
\end{pmatrix}  & 
\begin{pmatrix} 
\upsilon_{311} & \upsilon_{312} & \upsilon_{313}\\
\upsilon_{321} & \upsilon_{322} & \upsilon_{323}\\
\upsilon_{331} & \upsilon_{332} & \upsilon_{333}
\end{pmatrix} \\ 
\\
\begin{pmatrix} 
\upsilon_{411} & \upsilon_{412} & \upsilon_{413}\\
\upsilon_{421} & \upsilon_{422} & \upsilon_{423}\\
\upsilon_{431} & \upsilon_{432} & \upsilon_{433}
\end{pmatrix}  & 
\begin{pmatrix} 
\upsilon_{511} & \upsilon_{512} & \upsilon_{513}\\
\upsilon_{521} & \upsilon_{522} & \upsilon_{523}\\
\upsilon_{531} & \upsilon_{532} & \upsilon_{533}
\end{pmatrix}  & 
\begin{pmatrix} 
\upsilon_{611} & \upsilon_{612} & \upsilon_{613}\\
\upsilon_{621} & \upsilon_{622} & \upsilon_{623}\\
\upsilon_{631} & \upsilon_{632} & \upsilon_{633}
\end{pmatrix} \\
\\
\begin{pmatrix} 
\upsilon_{711} & \upsilon_{712} & \upsilon_{713}\\
\upsilon_{721} & \upsilon_{722} & \upsilon_{723}\\
\upsilon_{731} & \upsilon_{732} & \upsilon_{733}
\end{pmatrix}  & 
\begin{pmatrix} 
\upsilon_{811} & \upsilon_{812} & \upsilon_{813}\\
\upsilon_{821} & \upsilon_{822} & \upsilon_{823}\\
\upsilon_{831} & \upsilon_{832} & \upsilon_{833}
\end{pmatrix} & 
\begin{pmatrix}
\upsilon_{911} & \upsilon_{912} & \upsilon_{913}\\
\upsilon_{921} & \upsilon_{922} & \upsilon_{923}\\
\upsilon_{931} & \upsilon_{932} & \upsilon_{933}
\end{pmatrix}
\end{pmatrix}
\end{equation} 

By considering multiple isochronous surfaces it is possible to have complex predictive trajectory paths.

\section{Random segmentation surfaces for isochronous trajectories}
It is also possible to define a specific isochronous surface type and flag such segmentation surfaces that lead to changes in the trajectory path. To that extent, for a random segmentation surface $X$, the following considerations are made:
\begin{itemize}
\item there are $n$ values that characterize the segmentation surfaces $S_1, S_2,..,S_n$
\item there are $n$ surfaces dispersals $D^2(X_1), D^2(X_2), ...,D^2(X_n) $ that characterize the segmentation surfaces distributions
\item there are $n(n-1)$ isochronous correlations defined as:
\[ K_{ij}=V_{{iso}_{ij}}[(X_i-m_i)(X_j-m_j)] \] 
\end{itemize}

where:
\begin{itemize}
\item $X_i, X_j$ are the random segmentation surfaces
\item $m_i, m_j$ are the mean expected surface values
\item $V_{{iso}_{ij}}$ is the vehicle velocity isochronous matrix specific to the respective random segmentation surfaces
\end{itemize}

For situations when it is required to determine the distribution of a random segmentation surface event from the trajectory path, a particular case of the correlation can be used:
\[ D^2(X_i)=K_{ii}=V_{{iso}_{ij}}[(X_i-m_i)^2] \] 

For the complete trajectory $\Gamma$ it is possible to construct a correlations distribution matrix that describes the distribution of random segmentation surfaces $S_1,S_2,...,S_n$:

\[
\Gamma_{ij} = 
\begin{pmatrix}
K_{11} & K_{12} & ... & K_{1n}\\
K_{21} & K_{22} & ... & K_{2n}\\
...&...&...&...\\
K_{n1} & K_{n2} & ... & K_{nn}\\
\end{pmatrix}
\] 

Since the defined correlation allows the equality $K_{ij}=K_{ji}$, the  distribution matrix will also contain this equality and can be represented in the following way:

\[
\Gamma_{ij} = 
\begin{pmatrix}
K_{11} & K_{12} & ... & K_{1n}\\
       & K_{22} & ... & K_{2n}\\
 & &...&\\
 & & ... & K_{nn}\\
\end{pmatrix}
\]

As a consequence the isochronous trajectory is expanded to include the random segmentation surfaces and the correlation distribution matrix for them:

\begin{equation}  \label{eq:isochronous trajectory with random segmentation}
\Gamma_{t} = f\left((x_{t},y_{t},z_{t}),\begin{pmatrix}
K_{11} & K_{12} & ... & K_{1n}\\
       & K_{22} & ... & K_{2n}\\
 & &...&\\
 & & ... & K_{nn}\\
\end{pmatrix} \right) 
\end{equation}

\subsection{Domain of multiple random segmentation surfaces}
It is well known that a geometric mesh represents the discretization of the geometric domain into smaller and simpler shapes. For the two dimension, the triangles and quadrilaterals meshes are used. For the tree dimensions, the tetrahedral and hexahedral meshes are used as defined by \citet {key:3}.

	Since the trajectory path is a mesh of isochronous surfaces, based on previous recorded path data and the received node information, it is possible to detect path domains where movement obstacles might exist.
	
	Such an obstacle domain can be broken down into groups of segmentation surfaces with various sizes. The spacing between such surfaces can be identical with the isochronous surfaces space. The boundary for the obstacle domain will be:
	 
\[OBS = S_1 \cup S_2 \] 

where the $S_1$ and $S_2$ will have the following representation:
\[(S_1) \mapsto X= \zeta_1(Y,Z)\]
\[(S_2) \mapsto X= \zeta_2(Y,Z)\]  

\begin{figure}[!h]
\centering
{\includegraphics[width=0.7\textwidth]{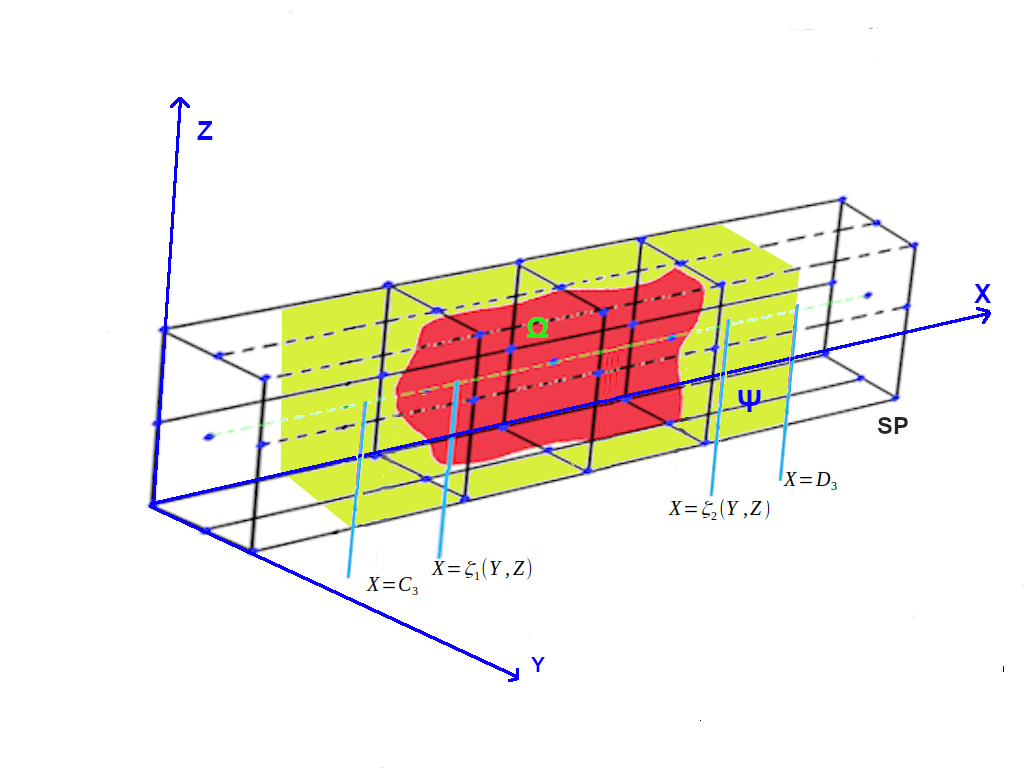}}
{\caption{Trajectory segmentation surfaces create a domain in the future trajectory path where risks exists.}}
\end{figure}

	The obstacle domain is named as $ \Omega $ and it is contained within the $ \Psi $ domain defined by the segmentation surface $X=C_3$ till $X=D_3$, where $C_3$ and $D_3$ are random surfaces that might or might not consider the time constraints that the isochronous surfaces follow.
	
	From this representation it is possible to define a function that defines the movement trough this section path SP. The resulting values can be found in the real numbers domain.
	
	\[ \hat f : SP \mapsto R  \]
	
\[ \hat f(x,y,z) = 
\left\{
	\begin{array}{ll}
		f(x,y,z)  & \mbox{, if } (x,y,z) \in \Omega \\
		0 & \mbox{, if } (x,y,z) \in  $ $ $$\Psi$ / $\Omega$$
	\end{array}
\right.
\]

From this it is possible to determine the function for the path section domain:

\[ \iiint\limits_{SP} \hat f(x,y,z) dxdydz =  \iiint\limits_\Omega \hat f(x,y,z) dxdydz + \iiint\limits_\Psi \hat f(x,y,z) dxdydz  \] 

\[ \iiint\limits_{SP} \hat f(x,y,z) dxdydz =  \iint\limits_{SP_{YZ}} \left( \int\limits_{C_3}^{D_3} \hat f(x,y,z) dx \right) dydz \] 

Now it is possible to define the complete $ \Omega $ domain and the $ \Psi $ domain following the X-axis:

\[ \int\limits_{C_3}^{D_3} \hat f(x,y,z) dx = \int\limits_{C_3}^{\zeta_1(Y,Z)} \hat f(x,y,z) dx + \int\limits_{\zeta_1(Y,Z)}^{\zeta_2(Y,Z)} \hat f(x,y,z) dx + \int\limits_{\zeta_2(Y,Z)}^{D_3} \hat f(x,y,z) dx\] 

	Or only considering the $\Omega$ domain:
	
\[ \int\limits_{C_3}^{D_3} \hat f(x,y,z) dx = 0 + \int\limits_{\zeta_1(Y,Z)}^{\zeta_2(Y,Z)} \hat f(x,y,z) dx + 0\]

Since the domain outside the $ \Omega $ domain is minimized to the 0 value, it can be considered that:

\[ \iiint\limits_{SP} \hat f(x,y,z) dxdydz =  \iiint\limits_\Omega f(x,y,z) dxdydz\]  

At the same time, if multiple path section domains are to be encountered, the trajectory using isochronous surfaces can be described as:

\[ \iiint\limits_{SP} f(x,y,z) dxdydz =  \sum\limits_{q=1}^n \iiint\limits_{\Omega_q} f(x,y,z) dxdydz\]  

From the defined isochronous surfaces it is possible to construct a trajectory mesh by using a mapping function. The coordinates of the nodes are known in the isochronous surface. The obtained mesh is then super imposed on the trajectory images.

As a consequence the isochronous trajectory is expanded to include the obstacle domain $\Omega$:

\begin{equation}  \label{eq:isochronous trajectory with obstacle domain}
\Gamma_{t} = f\left((x_{t},y_{t},z_{t}),\begin{pmatrix}
K_{11} & K_{12} & ... & K_{1n}\\
       & K_{22} & ... & K_{2n}\\
 & &...&\\
 & & ... & K_{nn}\\
\end{pmatrix} , \hat f(x,y,z) \right) 
\end{equation}

\subsection{Probability of segmentation surfaces}
Since a possible change in the trajectory is initialized by the segmentation surface, the distribution of the trajectory change can be modeled using the Laplace–Gauss distribution.

The density of repartition is described in \citet {key:7} as:

\[ f(x; m; \sigma) = \frac {1}{\sigma \sqrt 2\pi}e^{-\frac{1}{2}\frac{(x-m)^2}{\sigma^2}}\] 

where:
\begin{itemize}
\item $x$ real-valued random variable
\item $m$ maximum
\item $\sigma$ standard deviation
\end{itemize}

The probability that random segmentation surface $S_n$ takes a value between $(x_1, x_2)$ is given by:

\[ P(x_1 < S_n < x_2) = \int_{x_1}^{x_2} f(x; m; \sigma) dx = \frac {1}{\sigma \sqrt 2\pi}\int_{x_1}^{x_2} e^{-\frac{1}{2}\frac{(x-m)^2}{\sigma^2}}dx\] 

If we make the variable change $y=\frac{x-m}{\sigma}$ then $x=\sigma y + m$, $dx=\sigma dy$.
\begin{itemize}
\item $x= x_1$ then $y=\frac{x_1-m}{\sigma}$ 
\item $x= x_2$ then $y=\frac{x_2-m}{\sigma}$
\end{itemize}

\[ P(x_1 < S_n < x_2) = \frac {\sigma}{\sigma \sqrt 2\pi}\int_{\frac{x_1-m}{\sigma}}^{\frac{x_2-m}{\sigma}} e^{-\frac{y}{2}}dy = \frac {1}{\sqrt 2\pi} \left( \int_{-\infty}^{\frac{x_2-m}{\sigma}} e^{-\frac{y}{2}}dy - \int_{-\infty}^{\frac{x_1-m}{\sigma}} e^{-\frac{y}{2}}dy \right)\] 

If we consider the distribution function with the parameters 0 and 1, we can define the Laplace function $\Phi$:

\[ \Phi(x) = \int_{-\infty}^{x} f(y; 0; 1) dy = \frac {1}{\sqrt 2\pi}\int_{-\infty}^{x} e^{-\frac{y}{2}}dy\] 

and the probability that random segmentation surface $S_n$ takes a value between $(x_1, x_2)$ is given by:

\[ P(x_1 < S_n < x_2) = \Phi\left(\frac{x_2-m}{\sigma}\right) - \Phi\left(\frac{x_1-m}{\sigma}\right) \] 

As a consequence the isochronous trajectory is expanded to include the probability of random segmentation surfaces:

\begin{equation}  \label{eq:isochronous trajectory with probability}
\Gamma_{t} = f\left((x_{t},y_{t},z_{t}),\begin{pmatrix}
K_{11} & K_{12} & ... & K_{1n}\\
       & K_{22} & ... & K_{2n}\\
 & &...&\\
 & & ... & K_{nn}\\
\end{pmatrix} , \hat f(x,y,z), P(x_1 < S_n < x_2) \right) 
\end{equation}

\section{Isochronous trajectory chords}

Since it is possible to start from a node, follow an edge to the next node and then select another edge that is not adjacent to the first edge to continue to another edge, it is possible to define the concept of isochronous surface chords. They can be closed or open curves and can produce self-intersections.

\begin{theorem}
If one of the chords is disturbed, the disturbance is instantaneous transmitted throughout the complete $\Gamma_{t}$. 
\end{theorem}

By observing the amplitude of the chords disturbance it will be possible to build predictive paths for the remaining trajectory while at the same time updating the already known path.

As previously defined, the isochronous trajectory $\Gamma$, is composed of isochronous surfaces that interact with each other trough the chords.
The $\Gamma$ trajectory can be updated with new data from trajectories not yet travelled. The intent to change the already known $\Gamma$ trajectory with another $\Gamma$ trajectory, is captured by the "vibrations" of the isochronous surface chords defined as $\delta (t)$.

By analyzing the isochronous surface matrix of the velocity, acceleration and trajectory path updates it is possible to identify the isochronous surfaces that favour the appearance of chords vibrations.

\[ \delta (t) = \upsilon _{m,n,p}(t) +  \dot \upsilon _{m,n,p}(t) + \Omega _{m,n,p}(t)\] 

where:
\begin{itemize}
\item m,n,p are the indexes of the mini-nodes for each isochronous surface
\item $\upsilon _{m,n,p}(t)$ is the velocity
\item $\dot  \upsilon _{m,n,p}(t)$ is the acceleration
\item $\Omega _{m,n,p}(t)$ the obstacle domains 
\end{itemize}

By looking for the minimum $\delta (t)$ variation, it is possible to determine the complete trajectory profile of chords vibrations. 

\[ \Gamma (t) = f({\delta _{t_1}, \delta _{t_2}, ..., \delta _{t_n}}) \] 

where n is the number of isochronous surfaces with mini-nodes that have either chords vibrations or the probability of chords vibrations.

Since the trajectory to be traveled is not completely known, the chords vibrations have different propagation probabilities. For the path already traveled, the probability is $100\% $. For the path that is yet to be traveled, the chords vibrations propagation probability, will have the following form:

\[ \Gamma _{t} = f({\rho _{t_1}, \rho _{t_2}, ..., \rho _{t_n}}) \] 

where:
\begin{itemize}
\item $\Gamma _{t}$ is the trajectory
\item $\rho _{t}$ is the chords vibrations propagation probability
\item t is the time step
\end{itemize}

If the $\rho _{t}$ is higher then a certain threshold $\rho _{t_{min}}$, then the corresponding isochronous surface is selected as a segmentation surface.

As a consequence the isochronous trajectory is expanded to include the chords vibrations and the chords vibrations propagation probability:

\begin{equation}  \label{eq:isochronous trajectory with chords}
\Gamma_{t} = f\left((x_{t},y_{t},z_{t}),\begin{pmatrix}
K_{11} & K_{12} & ... & K_{1n}\\
       & K_{22} & ... & K_{2n}\\
 & &...&\\
 & & ... & K_{nn}\\
\end{pmatrix} , \hat f(x,y,z), P(x_1 < S_n < x_2), \delta _{t_n}, \rho _{t_n} \right) 
\end{equation}

\newpage

\section{Measurements}
	Instrumentation used in this study for the trajectory measurements consisted of a compass HDMM01 and a ArduinoUno.
The instruments used the factory calibration with no modifications. The sensor and ArduinoUno were fixed in a case that was placed on the vehicle floor. The 20 ms sample time between isochronous nodes was chosen in order to allow multiple readings within the target speed range of 0 to 50km/h. Also, the same driver was used to gather the data and the same trajectory was chosen to illustrate the method. Different hours and days where used to have a good sample of traffic influence on the trajectory isochronous patterns. 

	The measurement system is implemented with the sensor connected with a two wire (i2C) cable to the Arduino. The data from compass is the raw result of measured acceleration along a reference axis. No correction is done on the signal, no zero bias, zero scale factor or random error correction is applied.

\begin{table}[htb]
\begin{tabular}{ l | l | l | l | l }
\hline 
\multicolumn{5}{|c|}{Compass module}                                       \\ 
\hline \hline
\multicolumn{2}{|l|}{Type}                  & \multicolumn{3}{l|}{HDMM01,  Dual-axis Magnetic Sensor Module} \\
\hline
\multicolumn{2}{|l|}{Effective measurement range}                  & \multicolumn{3}{l|}{± 5 gausses} \\
\hline
\multicolumn{2}{|l|}{Sensitivity} & \multicolumn{3}{l|}{512counts/gauss @3.0V at 25°C} \\
\hline
\multicolumn{2}{|l|}{}                  & \multicolumn{3}{l|}{MIN: 461 counts/gauss} \\
\hline
\multicolumn{2}{|l|}{}                  & \multicolumn{3}{l|}{MAX: 563 counts/gauss} \\
\hline
\multicolumn{2}{|l|}{Supply Voltage}                  & \multicolumn{3}{l|}{2.7 … 5.25 V} \\
\hline
\multicolumn{2}{|l|}{Noise Density}                  & \multicolumn{3}{l|}{1 $\sim$ 25Hz, RMS, 600 \textmu gauss} \\
\hline
\multicolumn{2}{|l|}{Accuracy}                  & \multicolumn{3}{l|}{±5 deg} \\
\hline \hline
\multicolumn{5}{|c|}{Mechanical properties of the module}                                         \\
\hline 
\multicolumn{2}{|l|}{L × l × H}                  & \multicolumn{3}{l|}{16,20 × 12,30 × 2,25 (mm)} \\
\hline
\end{tabular}
\end{table}

	From these readings it was possible to derive the specific trajectory horizon of predictive isochronous paths. The trajectory graph consists of nodes, which are usually represented as circles, and edges, which are represented as line segments connecting the nodes. 

\begin{figure}[htb]
\centering
{\includegraphics[width=0.7\textwidth]{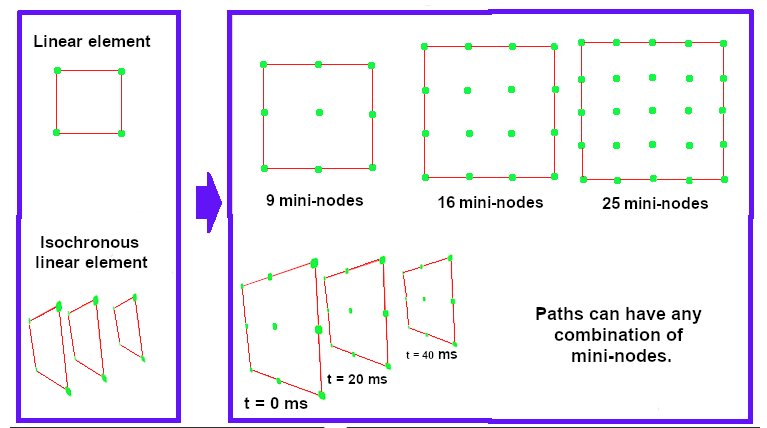}}
{\caption{Each isochronous surface can have multiple mini-nodes}}
\end{figure}

\begin{figure}[htb]
\centering
{\includegraphics[width=0.6\textwidth]{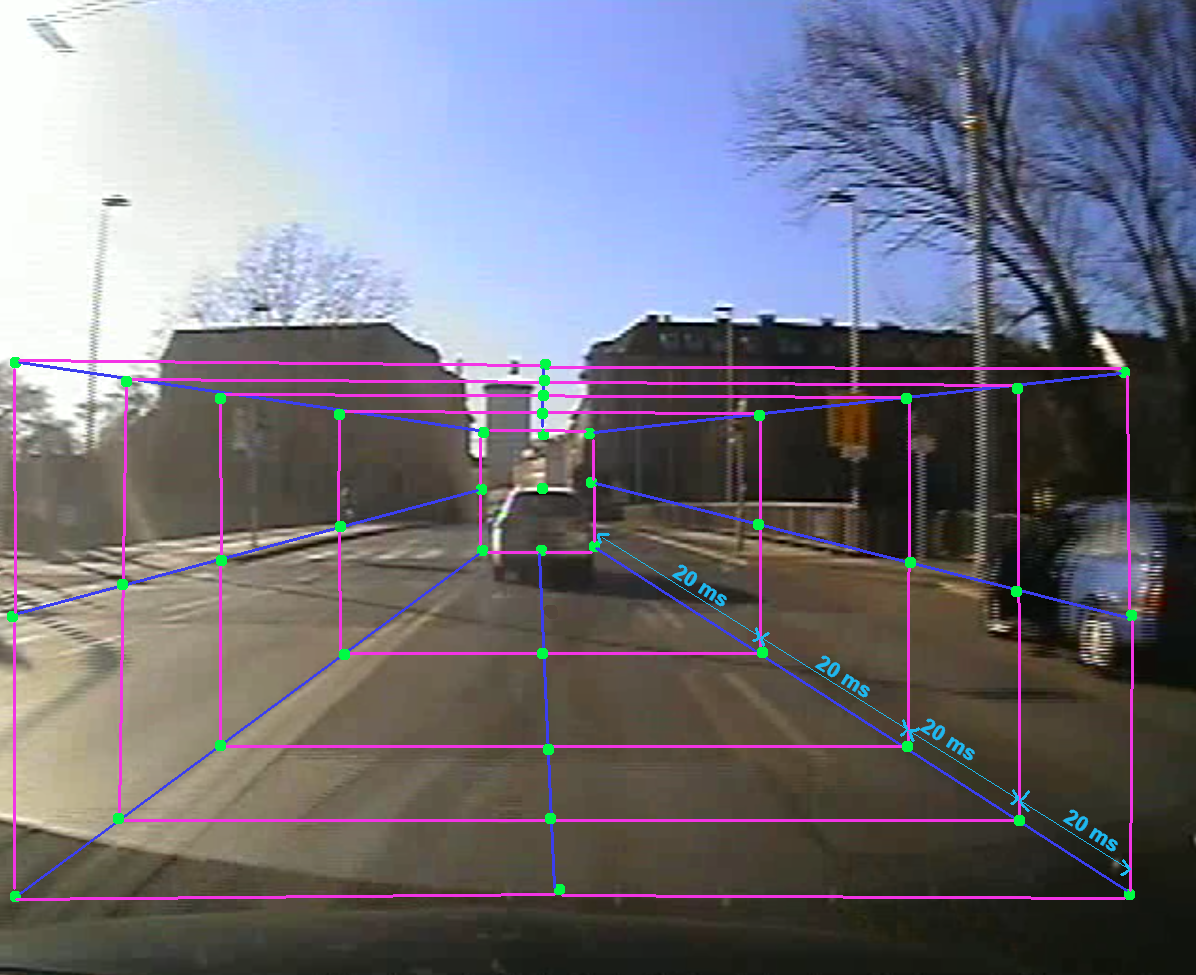}}
{\caption{Segment of vehicle path with isochronous surfaces}}
\end{figure}

	The forward direction can be derived from the corners of the isochronous surface or it can be derived from the center of the isochronous 9 mini-nodes representation. With the 16 mini-nodes representation, it is possible to have an increased precision for the trajectory prediction. Adding mini-nodes to each isochronous surface leads to improved prediction accuracy in situations like intersections or vehicle lane changing.	

	From the increased number of mini-nodes for each isochronous surface it is possible to reverse to less complex isochronous surface representation when the vehicle is approaching a stop light or an immediate standstill situation.

	Each node of the trajectory map represents a decision point, such as, “Must I make a left or a right turn?”. The corners of isochronous surface will indicate if the probability of a left or right turn decision is going to be increased or not. The predictive paths can be evaluated according to the benefits they offer, like reaching the destination with current vehicle energy reserves or improve operating parameters if the path adds a delay of minutes. From these paths, the optimal plan can be one that maximizes the vehicle state of operation. 
	
	Each mini-node and isochronous surface can give a certain probability that the vehicle will follow a specific path. A straight line trajectory can be represented with the green color, while an approaching vehicle path change can be represented from yellow to red. Previous compass data is used to provide predictive trajectory isochronous surfaces. Based on the compass data, the complexity of the isochronous surfaces can be reduced if there is a high probability that the vehicle will follow previous recorded paths. 

\newpage
	
\begin{figure}[!h]
\centering
{\includegraphics[width=0.9\textwidth]{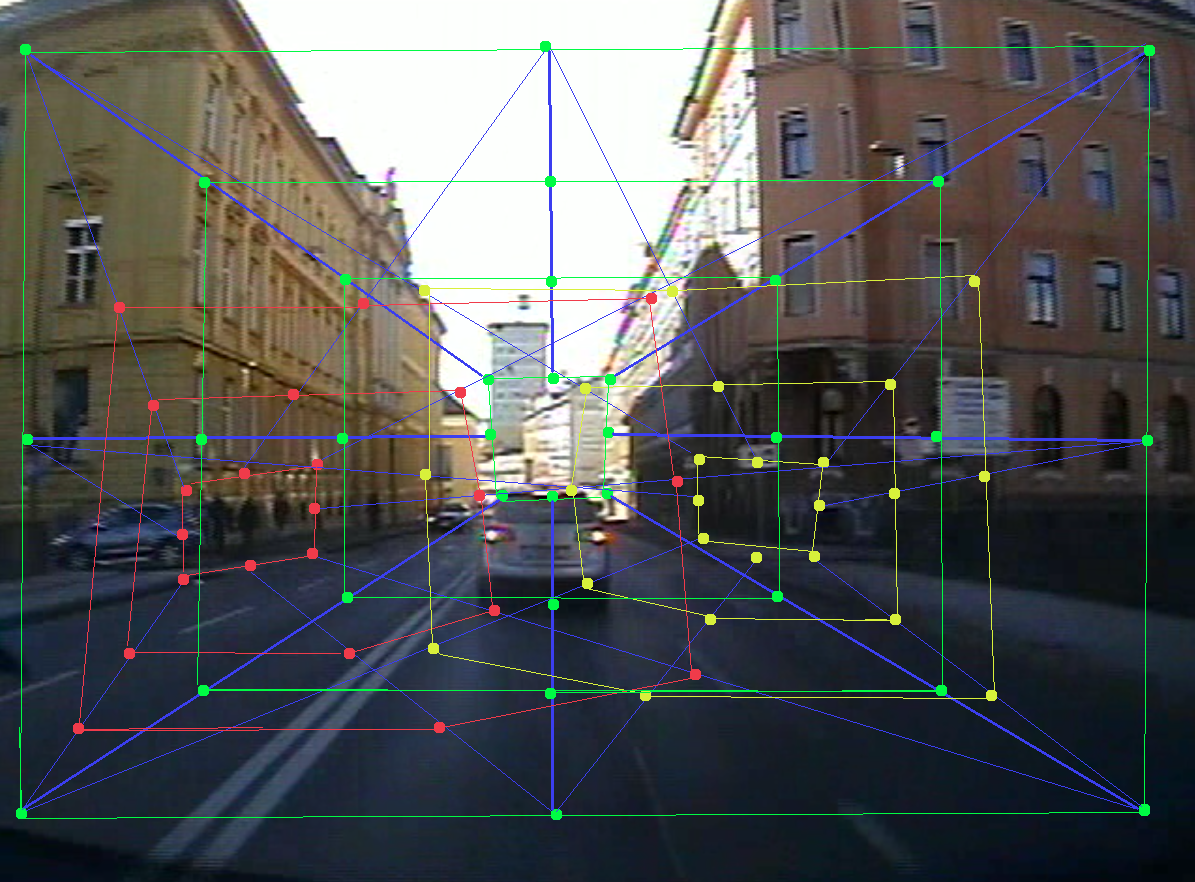}}
{\caption{Multiple isochronous trajectories.}{      Green (most likely), yellow (back-up path), red (not allowed)}\label{figure5}}
\end{figure}

	In figure \ref{figure5}, the vehicle will most likely follow the path no.1, as indicated by the color green in the isochronous trajectory surfaces. Nevertheless the path estimation can consider also path no.2 (red) and path no.3(yellow) as possible vehicle trajectories. The path no.2 (red) is a forbidden trajectory path. By having the video data and the compass data it is possible to validate the predictive trajectory isochronous surfaces.
	
	If the uncertainty level for a specific trajectory path reaches a certain level, it is possible to increase the number of mini-nodes for each isochronous surface. Thus, every small change in the sensor value can be captured and the prediction horizon for the vehicle trajectory can be increased.

The compass data can be used to determine also other variables that can characterize the vehicle behavior on the selected path:
\begin{itemize}
\item speed and shifting: slipping in and out of gears
\item acceleration and deceleration: the way a vehicle is maneuvered can influence how other vehicles react
\item changing lanes: it is possible to recognize lateral movements created by a vehicle and differentiate a left-lane change from a right-lane change
\end{itemize}

\begin{figure}[htbp]
\centering
{\includegraphics[width=0.8\textwidth]{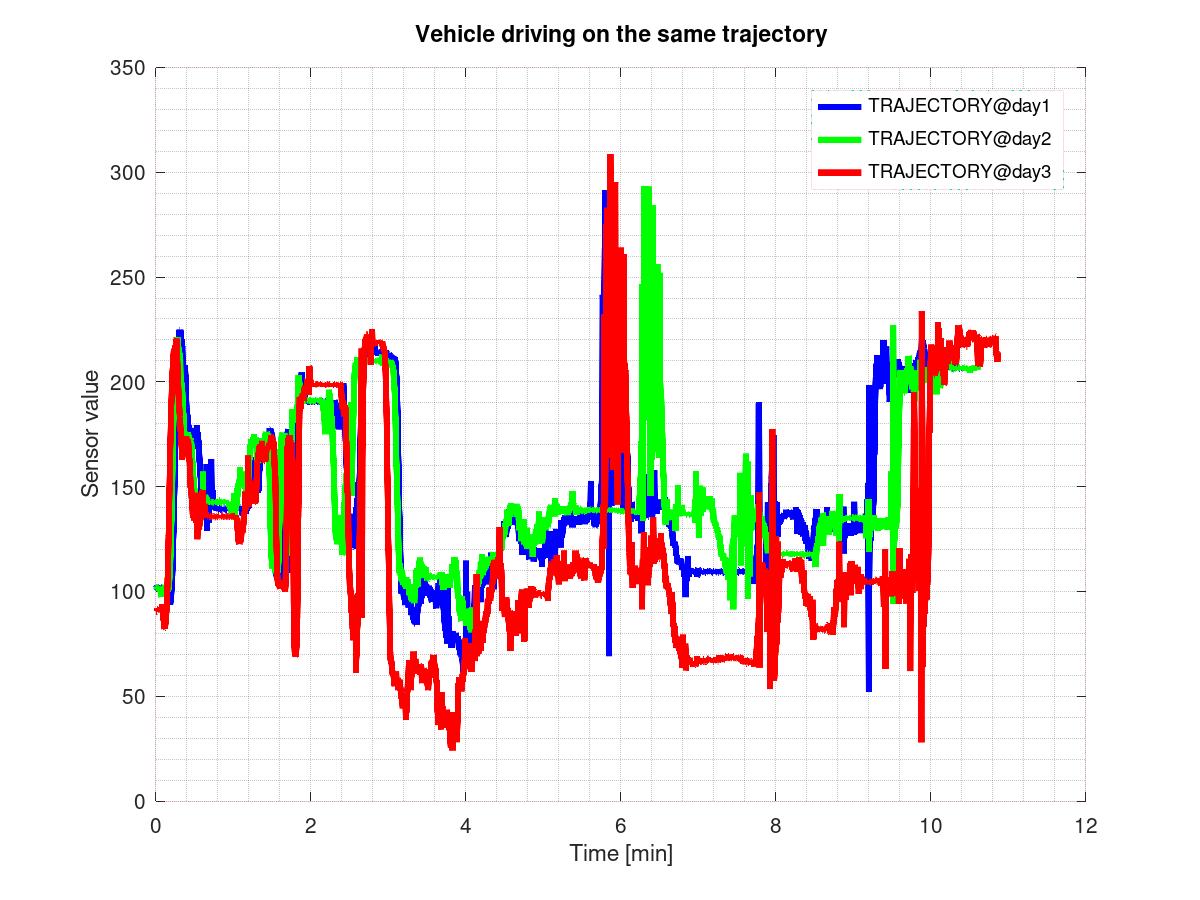}}
{\caption{Measurement results}}
\end{figure}

	The equation \ref{eq:isochronous trajectory with chords} provides the necessary information to calculate and predict the trajectory path $\Gamma_{t}$. For day 1, since the profile of trajectory path is not yet known, the information about possible obstacle domain $\Omega$ can only be received from the road infrastructure. In case of such obstacle domains $\Omega$ in day 1, the $\Gamma_{t}$ will be prepared to encounter them in day 2 and day 3. 
	In a similar way, the random segmentation surfaces, the correlation distribution matrix for them, the probability of random segmentation surfaces, the chords vibrations and the chords vibrations propagation probability are predicted with some degree of freedom in day 1. For day 2 and day 3 the accuracy of the $\Gamma_{t}$ predictability increases.

\section{Conclusions}
	In the present paper it was demonstrated the usage of predictive isochronus trajectories based on recorded compass sensor data. The measurement system is simple to construct, portable, cost efficient, low power consuming and reliable.
	
	By using predictive isochronus trajectories it is possible to determine vehicle path within a predefined time horizon. The mini-nodes that constitute the isochronus surfaces can provide great accuracy in predicting the desired trajectory.
	
	At the same time, energy savings can be obtain if the vehicle has a trajectory horizon estimation algorithm. Multiple energy systems can be synchronized to work together to achieve the best energy efficiency for the planned trajectory.

\newpage
\section{References}

\nocite{*}
\bibliography{article1.bib}

\begin{thebibliography}{13}
\providecommand{\natexlab}[1]{#1}
\providecommand{\url}[1]{\texttt{#1}}
\expandafter\ifx\csname urlstyle\endcsname\relax
  \providecommand{\doi}[1]{doi: #1}\else
  \providecommand{\doi}{doi: \begingroup \urlstyle{rm}\Url}\fi

\bibitem[Bern and Plassmann(2000)]{key:3}
Marshall Bern and Paul Plassmann.
\newblock Mesh generation.
\newblock In \emph{HANDBOOK OF COMPUTATIONAL GEOMETRY. ELSEVIER SCIENCE}, pages
  291--332, 2000.

\bibitem[Boskoff(1990)]{key:2}
V.~Boskoff.
\newblock \emph{Probleme practice de geometrie}.
\newblock Editura Tehnică, București, 1990.

\bibitem[Joshua and et~al.(2011)]{Autonomous}
Joseph Joshua and et~al.
\newblock A bayesian nonparametric approach to modeling motion patterns.
\newblock \emph{Autonomous Robots}, 2011.

\bibitem[Malgireddy et~al.(2013)Malgireddy, Nwogu, and
  Govindaraju]{JMLR:v14:malgireddy13a}
Manavender~R. Malgireddy, Ifeoma Nwogu, and Venu Govindaraju.
\newblock Language-motivated approaches to action recognition.
\newblock \emph{Journal of Machine Learning Research}, 14\penalty0
  (30):\penalty0 2189--2212, 2013.
\newblock URL \url{http://jmlr.org/papers/v14/malgireddy13a.html}.

\bibitem[Mayne et~al.(2000)Mayne, Rawlings, Rao, and Scokaert]{Automatica}
D.~Mayne, J.~Rawlings, C.~Rao, and P.~Scokaert.
\newblock Constrained model predictive control: Stability and optimality.
\newblock \emph{Automatica}, 2000.

\bibitem[Meier et~al.(2012)Meier, Theodorou, and Schaal]{key:6}
Franziska Meier, Evangelos Theodorou, and Stefan Schaal.
\newblock Movement segmentation and recognition for imitation learning.
\newblock In Neil~D. Lawrence and Mark Girolami, editors, \emph{Proceedings of
  Machine Learning Research}, volume~22, pages 761--769, La Palma, Canary
  Islands, 21--23 Apr 2012. PMLR.
\newblock URL \url{http://proceedings.mlr.press/v22/meier12.html}.

\bibitem[Moineagu(1976)]{key:7}
C.~Moineagu.
\newblock \emph{Statistica}.
\newblock Editura Didactică și Pedagogigă, București, 1976.

\bibitem[Peterson et~al.(2010)Peterson, Moore, Fintelman, and Hubbard]{bicycle}
Dale Peterson, Jason Moore, Danique Fintelman, and Mont Hubbard.
\newblock Low-power, modular, wireless dynamic measurement of bicycle motion.
\newblock \emph{Procedia Engineering}, 2:\penalty0 2949--2954, 06 2010.
\newblock \doi{10.1016/j.proeng.2010.04.093}.

\bibitem[POLLIN(2020)]{website:POLLIN}
POLLIN.
\newblock Kompassmodul hdmm01.
\newblock Available at \url{http://www.pollin.de/}, June 2020.

\bibitem[SAE(2020)]{website:SAE}
SAE.
\newblock Vehicle dynamics terminology j670-200801.
\newblock Available at \url{https://www.sae.org/standards/content/}, June 2020.

\bibitem[Sandu(2002)]{key:1}
M.~Sandu.
\newblock \emph{Mecanică teoretică}.
\newblock Editura Didactică și Pedagogigă, București, 2002.

\bibitem[Shao and Li(2015)]{key:5}
Zhanpeng Shao and Youfu Li.
\newblock Integral invariants for space motion trajectory matching and
  recognition.
\newblock \emph{Pattern Recogn.}, 48\penalty0 (8):\penalty0 2418–2432, August
  2015.
\newblock ISSN 0031-3203.
\newblock \doi{10.1016/j.patcog.2015.02.029}.

\bibitem[Sheffer et~al.(2007)Sheffer, Praun, and Rose]{key:4}
Alla Sheffer, Emil Praun, and Kenneth Rose.
\newblock Mesh parameterization methods and their applications.
\newblock \emph{Foundations and Trends® in Computer Graphics and Vision},
  2\penalty0 (2):\penalty0 105--171, 2007.
\newblock ISSN 1572-2740.
\newblock \doi{10.1561/0600000011}.
\newblock URL \url{http://dx.doi.org/10.1561/0600000011}.

\end{thebibliography}

\end{document}